\documentclass{article}

\usepackage{microtype}
\usepackage{graphicx}
\usepackage{subfigure}
\usepackage{booktabs} 
\usepackage{enumitem} 

\usepackage{hyperref}

\usepackage[accepted]{icml2025}

\usepackage{amsmath}
\usepackage{amssymb}
\usepackage{mathtools}
\usepackage{amsthm}

\usepackage[capitalize,noabbrev]{cleveref}

\usepackage{multirow}
\usepackage{multicol}
\usepackage{array}
\usepackage{soul}
\usepackage{tabularx}
\usepackage{float}
\theoremstyle{plain}
\newtheorem{theorem}{Theorem}[section]

\theoremstyle{definition}

\theoremstyle{remark}

\usepackage[textsize=tiny]{todonotes}

\icmltitlerunning{Direct Advantage Regression: Aligning LLMs with Online AI Reward}

\begin{document}

\twocolumn[
\icmltitle{Direct Advantage Regression: Aligning LLMs with Online AI Reward}



\icmlsetsymbol{equal}{*}

\begin{icmlauthorlist}
\icmlauthor{Li He}{unisyd,data61}
\icmlauthor{He Zhao}{data61}
\icmlauthor{Stephen Wan}{data61}
\icmlauthor{Dadong Wang}{data61}
\icmlauthor{Lina Yao}{data61}
\icmlauthor{Tongliang Liu}{unisyd}
\end{icmlauthorlist}

\icmlaffiliation{unisyd}{Sydney AI Centre, School of Computer Science, The University of Sydney}
\icmlaffiliation{data61}{CSIRO’s Data61}

\icmlcorrespondingauthor{Tongliang Liu}{tongliang.liu@sydney.edu.au}

\icmlkeywords{Machine Learning, ICML}

\vskip 0.3in
]



\printAffiliationsAndNotice{}  

\begin{abstract}
Online AI Feedback (OAIF) presents a promising alternative to Reinforcement Learning from Human Feedback (RLHF) by utilizing online AI preference in aligning language models (LLMs). However, the straightforward replacement of humans with AI deprives LLMs from learning more fine-grained AI supervision beyond binary signals. In this paper, we propose Direct Advantage Regression (DAR), a simple alignment algorithm using online AI reward to optimize policy improvement through weighted supervised fine-tuning. As an RL-free approach, DAR maintains theoretical consistency with online RLHF pipelines while significantly reducing implementation complexity and improving learning efficiency. Our empirical results underscore that AI reward is a better form of AI supervision consistently achieving higher human-AI agreement as opposed to AI preference. Additionally, evaluations using GPT-4-Turbo and MT-bench show that DAR outperforms both OAIF and online RLHF baselines.
\end{abstract}

\section{Introduction}
Large language models (LLMs) \cite{brown2020languagemodelsfewshotlearners, bubeck2023sparksartificialgeneralintelligence} have demonstrated their capability to replace human supervision in various tasks, leading to higher training efficiency and lower deployment costs \cite{burns2023weaktostronggeneralizationelicitingstrong, cui2024ultrafeedbackboostinglanguagemodels}. A vivid example of this is  Online AI Feedback (OAIF) \cite{guo2024directlanguagemodelalignment}, which aims to fine-tune LLMs using pairwise preference labels generated by AI annotators in an online learning setting. OAIF shares a similar training framework as Reinforcement Learning from AI Feedback (RLAIF) \cite{bai2022constitutionalaiharmlessnessai, lee2024rlaifvsrlhfscaling}, but introduces a key modification: an online iterative preference learning process with on-policy data generation. More specifically, the learning policy in OAIF learns directly from preference labels between pairs of on-policy responses collected in each online iteration. This results in a minimized distributional shift between the preference dataset and the learning policy, and eventually enables learning a policy outperforming RLAIF and Reinforcement Learning from Human Feedback (RLHF).

\begin{figure*}[t]
\vspace{0mm}
\begin{center}
\centerline{{\includegraphics[width=2\columnwidth, clip, trim = 5mm 30mm 5mm 50mm]{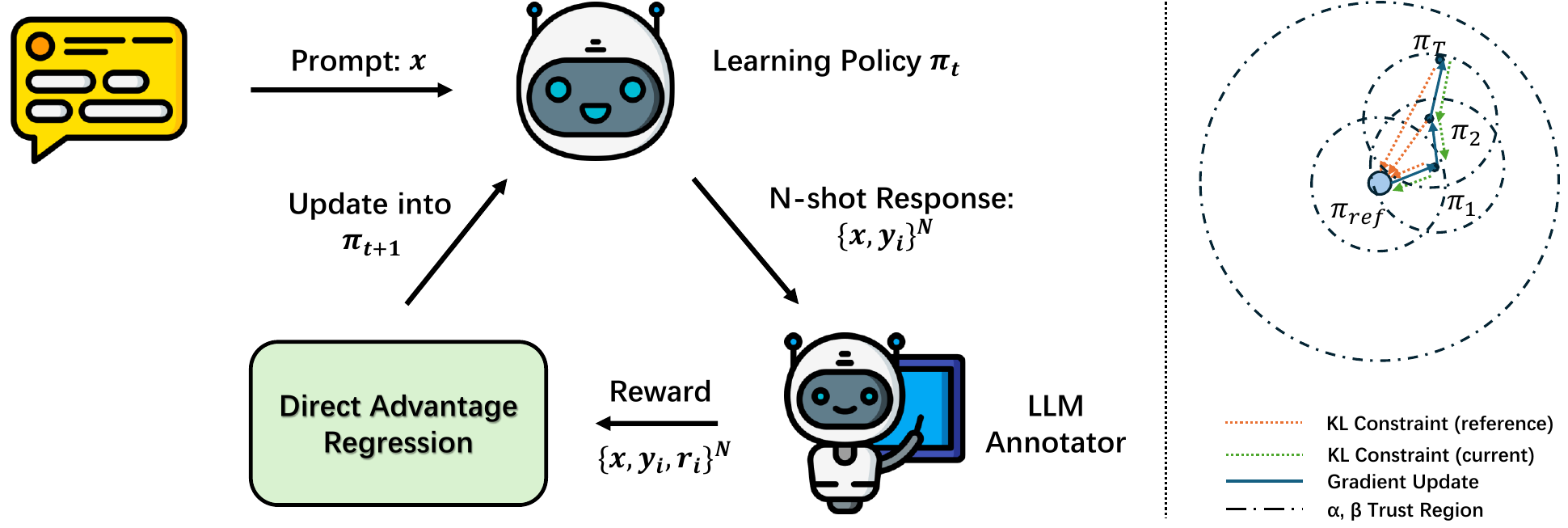}}}
\caption{\textbf{Direct Advantage Regression with Online AI Reward.} 
(Left) Using the reward labels provided by the LLM annotator, DAR increases the likelihood of each n-shot responses based on the calculated regression weight, so that the response of higher quality will have a higher probability to be sampled in the next iteration. 
(Right) The dual-constraint optimization objective of DAR: 1) the reference regularization prevents reward over-optimization, 2) the current sampling regularization ensures stable gradient updates in each iteration.
}
\label{fig:dar}
\end{center}
\vspace{-5mm}
\end{figure*}

However, the OAIF framework requiring LLM supervision in the form of binary preferences certainly bottlenecks a broaden deployment of LLM annotators. While LLM annotators are able to give more expressive supervision signals, such as preference margins and equivalences, the OAIF-like preference learning frameworks discard this information \cite{chen2024bootstrappinglanguagemodelsdpo, chen2024optuneefficientonlinepreference, chen2024costeffectiveproxyrewardmodel, pang2024iterativereasoningpreferenceoptimization,yuan2024selfrewardinglanguagemodels}, leading to a loss of fine-grained task understanding. 

Moreover, the off-policy Direct Alignment from Preference (DAP) methods \cite{ azar2023generaltheoreticalparadigmunderstand, zhao2023slichfsequencelikelihoodcalibration, rafailov2024directpreferenceoptimizationlanguage} used in OAIF have not been adequately adapted for online learning scenarios, making them suboptimal for iterative online alignment.

In this paper, we show how to more efficiently leverage online AI supervision to align LLMs meanwhile avoiding complex implementations of RL. We propose Direct Advantage Regression (DAR), an online alignment algorithm that optimizes policy improvement through an expectation-maximization framework leveraging online AI reward. Via optimizing a weighted supervised fine-tuning (SFT) loss, DAR iteratively increases the log probability of sampled responses proportional to their advantage, meanwhile effectively avoiding reward hacking \cite{ziegler2020finetuninglanguagemodelshuman, bai2022traininghelpfulharmlessassistant, ouyang2022traininglanguagemodelsfollow, stiennon2022learningsummarizehumanfeedback} and ensuring stable policy improvement \cite{schulman2017trustregionpolicyoptimization, peng2019advantageweightedregressionsimplescalable}. In contrast to the DAP methods, DAR is a proper on-policy learnring approach offers better online convergence properties and enhanced learning efficiency. On the other hand, DAR offers a RL-free alternative with a simpler implementation compared to conventional online RLHF methods such as PPO \cite{schulman2017proximalpolicyoptimizationalgorithms}.  

We summarize our contributions as follows:
\begin{itemize}[leftmargin=0.4cm,topsep=-2pt]
    \item We perform a head-to-head comparison of AI reward and AI preference using a wide range of models as AI annotators. Our results show that AI reward consistently reaches a higher level of agreement with human preferences.
    \item We present the DAR algorithm with detailed mathematical derivations, and empirically demonstrate its advantage over online RLHF algorithms and DAP algorithms in the online setting of direct AI alignment.
    \item We apply DAR in online RLHF using a pre-trained reward model. Evaluations, judged by GPT-4, show that the aligned model outperforms the baselines produced by online RLHF methods and a list of open-source LLMs.
    \item We conduct comprehensive ablation studies that bridges empirical observations with theoretical understanding, facilitating broader downstream applications of DAR.
\end{itemize}

\section{Related Work}
\subsection{Learning from LLMs}
Learning from LLMs is a prevailing paradigm and has proven to be both effective and cost-efficient. \citet{bai2022constitutionalaiharmlessnessai} first proposed the concept of RLAIF, where an LLM is used to provide response refinement and preference feedback based on a set of human-written principles to facilitate AI safety alignment. \citet{lee2024rlaifvsrlhfscaling} were the first to conduct a thorough comparison of RLHF versus RLAIF, where the reward model is trained separately using solely human preferences or AI preferences. Their findings indicated that the models via RLAIF were more preferred, while the cost of RLAIF is ten times cheaper. Moreover, a significant line of research investigates self-learning LLMs, where the student and teacher models are identical. These self-aligning LLMs learn without external supervision, utilizing either self-generated high-quality datasets \cite{huang2022largelanguagemodelsselfimprove, wang2023selfinstructaligninglanguagemodels, li2024selfalignmentinstructionbacktranslation} or self-annotated pairwise preference labels \cite{chen2024grathgradualselftruthifyinglarge, yuan2024selfrewardinglanguagemodels}. Unlike previously discussed methods, we align LLMs using AI-generated scalar reward labels within a framework similar to direct-RLAIF proposed by \citet{lee2024rlaifvsrlhfscaling}. This online alignment setup fully embraces more expressive AI supervision by directly employing LLMs as reward models.

\subsection{Online Alignment Algorithms}
The dominant alignment algorithm in RLHF has been PPO \cite{schulman2017proximalpolicyoptimizationalgorithms}, which has demonstrated notable stability and effectiveness as an online policy-gradient method. However, the implementation complexity of PPO, particularly the requirement for an additional value model, has motivated the search for simpler alternatives. In response, REINFORCE Leave-One-Out (RLOO) \cite{kool2019buy, ahmadian2024basics} introduced a streamlined approach that utilizes Monte-Carlo sampling for baseline value estimation, eliminating the need for fitting a separate value model. Moreover, substantial research has focused on online DAP methods through incorporating online feedback mechanisms while keeping the algorithm untouched. The online preference labels are generated using a reward model \cite{chen2024optuneefficientonlinepreference, xu2024dposuperiorppollm}, direct human feedback \cite{xiong2024iterativepreferencelearninghuman}, or LLMs annotators\cite{guo2024directlanguagemodelalignment, qi2024onlinedpoonlinedirect}. In contrast to existing approaches, DAR explicitly implements KL regularization between the learning policy and sampling policy, and our work represents another endeavor to explore simpler alternatives to PPO.

\section{Preliminaries}

\subsection{RL Fine-tuning}
Given an LLM to be aligned $\pi_\theta$, a prompt dataset $\mathcal{D}(x)$ and a reward model $r$, online RL fine-tuning \cite{ziegler2020finetuninglanguagemodelshuman, bai2022traininghelpfulharmlessassistant, ouyang2022traininglanguagemodelsfollow, stiennon2022learningsummarizehumanfeedback} aims to optimize a reward maximizing objective using online sampled response $y$ with an extra reference regularization term: \begin{equation}\begin{split}
    \mathcal{J}_\text{RLHF}(\pi_\theta;\pi_\text{ref}) = \mathop{\mathrm{max}}_{\pi_\theta}\mathbb{E}_{x \sim \mathcal{D}(x),y \sim \pi_\theta(y|x)} \Bigl[ r(x,y) \Bigr] \, & \\
    -\alpha \mathbb{D}_{\textrm{KL}}  \Bigl[ \pi_\theta(y|x) \parallel \pi_\text{ref}(y|x) \Bigr] &,
\end{split}\label{eq:llm_align}
\end{equation}
where $\alpha$ is a coefficient controlling the KL divergence between $\pi_\theta$ and a reference policy $\pi_\text{ref}$. The purpose of the KL regularization is to preserve previously acquired knowledge and also mitigate the issue of reward hacking. To establish an effective and proper regularization target, the reference policy is usually initialized to $\pi^\text{SFT}$, a model that has been supervised fine-tuned on a high-quality dataset for downstream tasks. Hence, in this scenario, the reference policy $\pi_\text{ref}$, the supervised fine-tuned policy $\pi^\text{SFT}$, and the initialization policy $\pi_{t=0}$ are identical.

\subsection{Advantage Weighted Regression}
Advantage Weighted Regression (AWR) \cite{peng2019advantageweightedregressionsimplescalable} iteratively improves a policy $\pi_\theta$ by maximizing an expectation of policy improvement over the current policy $\pi_t$:
\begin{equation}
\mathcal{J}_\text{AWR}(\pi_{\theta})=\mathop{\mathrm{max}}_{\pi_\theta}\mathbb{E}_{x \sim d_{\pi_\theta}(x),y \sim \pi_\theta(y|x)} \Bigl[ A(x,y) \Bigr],
\label{eq:awr_obj_1}
\end{equation}
where $d_{\pi_\theta}$ is the induced input distribution for $\pi_\theta$ \cite{sutton1998introduction}, $A(x, y)=r(x,y)-V^{\pi_t}(x)$ is the advantage function reflecting the expected improvement with respect to $\pi_t$ and $V^{\pi_t}(x)$ is the value function, representing the expected reward for $\pi_t$. In the context of auto-regressive generation employed by transformer-based LLMs \cite{radford2018improving, vaswani2023attentionneed}, the input distribution exhibits clear model dependence. We abuse the notation $d_{\pi_\theta}$ to represent the distribution of input token sequences, encompassing both the initial prompt and previously generated outputs by $\pi_\theta $, for simplicity.

To remove the dependency on dynamically estimating $d_{\pi_\theta}$, they instead use $d_{\pi_t}$ to replace $d_{\pi_\theta}$, and formulate a constrained policy optimization objective  as an approximation to \cref{eq:awr_obj_1}:
\begin{equation}\begin{split}
\mathcal{J}_\text{AWR}(\pi_{\theta};\pi_t)=\mathop{\mathrm{max}}_{\pi_\theta}\mathbb{E}_{x \sim d_{\pi_t}(x),y \sim \pi_\theta(y|x)} \Bigl[ A(x,y) \Bigr] \, & \\
    - \beta \mathbb{D}_{\textrm{KL}} \Bigl[ \pi_\theta(y|x) \parallel \pi_t(y|x) \Bigr] &,
\label{eq:awr_obj_2}
\end{split}
\end{equation}
where $\beta$ is a positive regularization coefficient ensuring such an approximation is valid. And finally, AWR solves this problem via the expectation maximization framework and iteratively finds a new policy $\pi_{t+1}$ with a higher reward expectation using a supervised regression loss:
\begin{equation}\begin{split}
\pi_{t+1} = \mathop{\mathrm{\arg  \max}}_{\pi_\theta} \mathbb{E}_{x \sim d_{\pi_t}(x), y \sim \pi_{t}(y|x)} \qquad \qquad \qquad &\\
  \left[ \mathrm{log} \pi_\theta(y | x) \mathrm{exp}\left( \frac{1}{\beta} A(x,y) \right)\right].&
\end{split}
\end{equation}

\begin{figure}[t]
\vspace{0mm}
\begin{center}
\centerline{{\includegraphics[width=\columnwidth, clip, trim = 0mm 90mm 172mm 0mm]{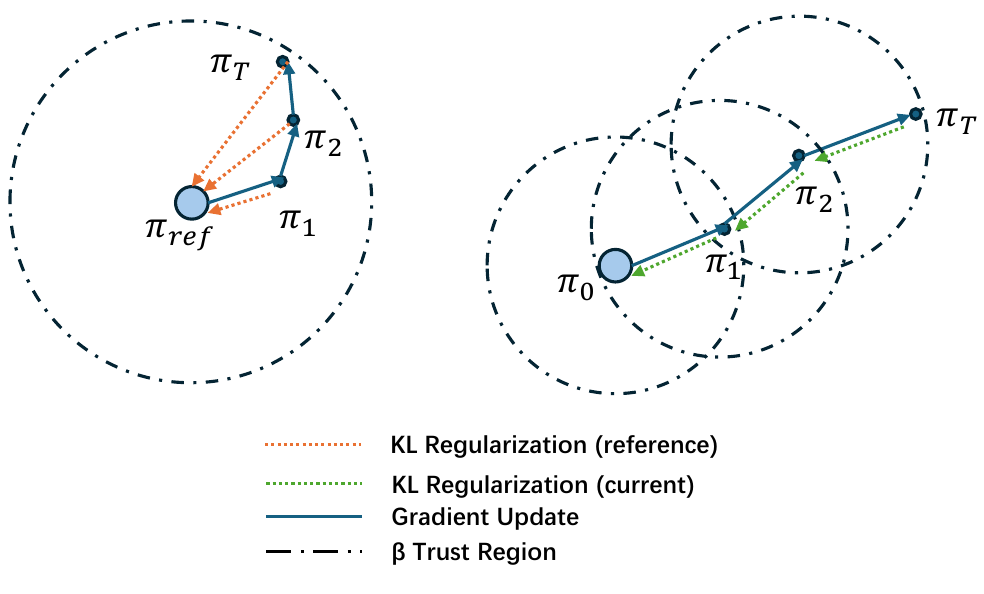}}}
\vspace{-2mm}
\caption{\textbf{Contrasting KL regularization approaches in RL fine-tuning and on-policy RL.}
(Left) RL fine-tuning employs a fixed reference policy to mitigate reward hacking.
(Right) On-policy RL methods (including regression-based) regularize with respect to the current sampling policy to ensure monotonic policy improvement.
}
\label{fig:kl_constrast_rlhf_on_policy_rl}
\end{center}
\vspace{-10mm}
\end{figure}

\section{Direct Advantage Regression}
This section presents a streamlined optimization approach that significantly simplifies implementation compared to conventional RLHF pipelines. Our formulation incorporates dual regularization targets: a reference policy and the current sampling policy. As shown in \cref{fig:dar} (Right), this objective aligns with conventional online RLHF pipelines, which combine reference-shaped rewards with PPO's on-policy improvement guarantees shown in \cref{fig:kl_constrast_rlhf_on_policy_rl}.

Through a regression-based iterative learning framework, we transform this dual-constrained RL objective into a weighted SFT loss. The weights capture two critical dimensions: (1) the advantage, which quantifies response quality, and (2) the regularization discount, which reflects model confidence in sampled responses as governed by the dual KL regularization constraints. This simplified implementation maintains the theoretical guarantees of conventional RLHF while reducing both computational overhead and implementation complexity.

\subsection{Derivation}
To adapt the iterative regression-based framework onto the task of LLM alignment, we introduce the reference regularization of \cref{eq:llm_align} into \cref{eq:awr_obj_2} and formulate a dual-constrained policy improvement objective:
\begin{equation}\begin{split}
&\mathcal{J}_\text{DAR}(\pi_\theta; \pi_\text{ref}, \pi_t) = \mathop{\mathrm{max}}_{\pi_\theta} \mathbb{E}_{x \sim d_{\pi_t}(x), y \sim \pi_\theta(y|x)} [A(x,y)]   \\
&- \alpha \mathbb{D}_{\textrm{KL}} \Bigl[ \pi_\theta(y|x) \parallel \pi_\text{ref}(y|x) \Bigr] - \beta \mathbb{D}_{\textrm{KL}} \Bigl[ \pi_\theta(y|x) \parallel  \pi_t(y|x)\Bigr].
\end{split}\label{eq:dar_obj}
\end{equation}
Following previous works, we derive the following Theorem with detailed proof in \cref{sec:proof}:
\begin{theorem}
Under mild assumption, given a dual-constrained advantage (or reward) maximization objective such as the one in \cref{eq:dar_obj}, with two KL coefficients being strictly positive, there exists a solution to the problem:
\begin{equation*}
     \pi^*=\frac{1}{Z(x)}  \pi_{\textnormal{ref}}(y|x)^{\frac{\alpha}{\alpha+\beta}} \pi_t(y|x)^{\frac{\beta}{\alpha+\beta}}   \exp\left(\frac{A(x,y)}{\alpha+\beta}\right),
\end{equation*}
where $ Z(x)=\sum\limits_{y} \pi_{\textnormal{ref}}(y|x)^{\frac{\alpha}{\alpha+\beta}}\pi_t(y|x)^{\frac{\beta}{\alpha+\beta}}\exp\left(\frac{A(x,y)}{\alpha+\beta}\right)$ is the partition function. \label{thm:dual_constrained_theorem}
\end{theorem}
We can now obtain an improved policy $\pi_\theta$ parameterized with $\theta$ by minimizing the KL-divergence between itself and the optimal policy $\pi^*$ defined in \cref{thm:dual_constrained_theorem}:
\begin{equation}
\mathop{\mathrm{min}}_{\pi_\theta} \mathbb{E}_{x \sim d_{\pi_t}(x)} \mathbb{D}_{\textrm{KL}} \Bigl[ \pi^*(\cdot|x) \parallel \pi_\theta(\cdot|x) \Bigr].
\label{eq:KL_obj}
\end{equation}
Through substitution and mathematical reductions shown in Appendix \ref{DAR_derivation}, we ultimately transform \cref{eq:KL_obj} into the following optimization objective for solving the iterative policy regression problem:
\begin{equation}\begin{split}
\pi_{t+1}=\mathop{\mathrm{arg \ max}}_{\pi_\theta} \mathbb{E}_{(x,y) \sim \mathcal{D}_{\pi_t}} \qquad \qquad \qquad \qquad \qquad &\\
\left[  \left(\frac{\pi_{\text{ref}}(y|x)}{\pi_t(y|x)}\right)^{\frac{\alpha}{\alpha+\beta}}  \exp\left(\frac{A(x,y)}{\alpha+\beta}\right) \log\pi_{\theta}(y|x) \right],&
\end{split}\label{eq:dar_loss}
\end{equation}
where $\mathcal{D}_{\pi_t}$ is the online dataset consists of prompt-response pairs collected by $\pi_t$.
\subsection{Gradient Analysis}
The iterative policy search loss in DAR manifests through weighted supervised fine-tuning based on prompt-response pairs that the model generates auto-regressively at each iteration. We calculate the gradient of \cref{eq:dar_loss} with respect to the parameter $\theta$: 
\begin{multline*}\label{eq:gradient}
    \nabla_\theta \mathcal{L}_\text{DAR}(\pi_\theta;\pi_\text{ref}, \pi_t) = -\mathbb{E}_{(x, y) \sim \mathcal{D}_{\pi_t}} \\ \bigg[\underbrace{ \left(\frac{\pi_{\text{ref}}(y|x)}{\pi_t(y|x)}\right)^{\frac{\alpha}{\alpha+\beta}}}_{\textcolor{blue}{\textnormal{Regularization Weight}}} \, \underbrace{\exp\left(\frac{A(x,y)}{\alpha+\beta}\right)}_{\textcolor{red}{\textnormal{Advantage Weight}}}
    \underbrace{\raisebox{-2.1ex}{\phantom{}}\nabla_\theta\log \pi_\theta(y\mid x)}_\text{SFT: increase likelihood of y} \bigg].
\end{multline*}
The gradient increases the log probability of each response proportional to the product of 1) \textcolor{blue}{Regularization Weight}: a discount factor that penalizes responses proportional to their divergence away from the reference distribution, and 2) \textcolor{red}{Advantage Weight}: a reward signal that is higher for a higher expectation in policy improvement. Besides, in terms of the impact of KL coefficients on the gradient, the sum of coefficients $\alpha+\beta$ serves as the scaling temperature for \textcolor{red}{Advantage Weight}, while a higher ratio of $\alpha$ over the sum leads to a more conservative \textcolor{blue}{Regularization Weight}. In other words, the gradient update drives the learning policy towards a distribution of higher policy improvement, with a soft regularization considering a combination of targets: 1) a static reference policy, and 2) a dynamic sampling policy, with $\alpha$ and $\beta$ regulating the total and relative strengths. 

\begin{algorithm}[h]
   \caption{Direct Advantage Regression}
   \label{alg:dar}
\begin{algorithmic}
    \STATE {\bfseries Input:} prompt dataset $\mathcal{D}(x)$, reference model $\pi_{\text{ref}}$, reward model $r$, training steps $T$, regularization coefficients $\alpha, \beta$, Monte-Carlo sampling size $K$, clip threshold $w_{\textnormal{clip}}$
    \STATE Initialize $\pi_\theta = \pi_{\text{ref}}, \pi_{t=0} = \pi_{\text{ref}}$.
    \FOR{$t=0$ {\bfseries to} $T-1$}
    \STATE Sample prompt $x$ from $\mathcal{D}(x)$
    \STATE Sample K-shot responses $\{y_{i}\}^{K}_{i=1}$ from $\pi_t(\cdot|x)$
    \STATE Calculate Advantage \\  \qquad$A(x, y_i) =r(x, y_i) - \frac{1}{K} \sum_{i=1}^K r(x,y_i)$
    \STATE Apply Advantage Normalization \\  \qquad $A_{\textnormal{norm}}(x, y_i) = [ A(x, y_i) - \mu_A ] / \sigma_{A}$
    \STATE Calculate Advantage and Regularization Weight \\  \qquad $\textcolor{red}{w_{\textnormal{adv}}^i}= \exp\left(\frac{1}{\alpha+\beta}A_{\textnormal{norm}}(x,y_i)\right) $\\  \qquad $\textcolor{blue}{w_{\textnormal{reg}}^i}= \left(\frac{\pi_{\text{ref}}(y_i|x)}{\pi_t(y_i|x)}\right)^{\frac{\alpha}{\alpha+\beta}}$
    \STATE Apply Weight Clip \\ \qquad ${w_{\textnormal{DAR}}^i}= \min(\textcolor{blue}{w_{\textnormal{reg}}^i} \cdot \textcolor{red}{w_{\textnormal{adv}}^i}, w_{\textnormal{clip}}) $
    \STATE Update $\theta_t$ into $\theta_{t+1}$ using $\nabla_\theta \mathcal{L}_\text{DAR}(\pi_{\theta})=$ \\ \qquad $-\frac{1}{K}\sum_{i=1}^K [ {w_{\textnormal{DAR}}^i} \nabla_\theta\log \pi_\theta(y_i\mid x)]$
   \STATE Let $\pi_{t+1} = \pi_{\theta_{t+1}}$
   \ENDFOR
\end{algorithmic}
\end{algorithm}

\begin{table*}[t]
\caption{Human-AI agreement of AI reward versus AI preference based on a 1k subset of test set of TL;DR, Helpfulness, and Harmlessness using LLM annotators from Qwen2, Llama-3, Mistral, Gemma-2 and GPT-4. To mitigate the positional bias, the agreement for AI preference is averaged over (chosen vs. rejected) and (rejected vs. chosen). The results for Llama-3 on Harmlessness is not available due a large amount of invalid judgments generated due to the triggered security mechanism.}
\label{tab:reward_vs_preference}
\vskip 0.15in
\begin{center}
\begin{small}
\begin{sc}
\begin{tabular}{llcccccccr}
\toprule
\multirow{2}{*}{Model}                   &  \multirow{2}{*}{Version}        & \multicolumn{3}{c}{AI Reward}     & \multicolumn{3}{c}{AI Preference}  & \text{Reward over}         \\
\cmidrule(lr){3-5} \cmidrule(lr){6-8}
& & \scriptsize{TL;DR} & \scriptsize{Helpful} & \scriptsize{Harmless} & \scriptsize{TL;DR} & \scriptsize{Helpful} & \scriptsize{Harmless} & {Preference?}\\
\midrule
\multirow{2}{*}{Qwen2}  
& 72B-Instruct  
& 74.97\%   & 73.25\%   & 73.89\%   & 71.35\%   & 71.15\%   & 67.19\%   & \multirow{2}{*}{\textcolor[RGB]{0,100,0}{$\surd$}}\\
& 7B-Instruct   
& 67.32\%   & 72.10\%	& 61.86\%   & 65.74\%	& 66.42\%	& 61.40\%  \\
\midrule
{\multirow{2}{*}{Lllam-3}}
& 70B-Instruct
& 73.70\%	& 73.07\%	& \multirow{2}{*}{N/A}  & 58.13\%	& 69.82\%	& \multirow{2}{*}{N/A}       & \multirow{2}{*}{\textcolor[RGB]{0,100,0}{$\surd$}}    \\
& 8B-Instruct   
& 61.15\%	& 70.16\%   &                       & 60.95\%	& 66.03\%       \\
\midrule
\multirow{2}{*}{Mistral}
& 8x7B-Instruct 
& 75.86\%	& 72.35\%	& 72.13\%   & 67.76\%	& 68.59\%	& 51.22\%   & \multirow{2}{*}{\textcolor[RGB]{0,100,0}{$\surd$}}    \\
& 7B-Instruct   
& 69.85\%	& 68.09\%	& 70.05\%   & 64.55\%	& 67.01\%	& 63.50\%   \\
\midrule
\multirow{2}{*}{Gemma-2}
& 27b-it        
& 74.84\%	& 72.44\%	& 74.44\%   & 68.45\%   & 67.37\%   & 68.93\%   & \multirow{2}{*}{\textcolor[RGB]{0,100,0}{$\surd$}}    \\                
& 9b-it         
& 74.87\%	& 70.66\%	& 72.71\%   & 67.36\%	& 66.73\%	& 67.43\%   \\
\midrule
Llama-3.1               
& 405B-Instruct 
& 79.32\%	& 74.34\%	& 81.58\%   & 72.76\%	& 68.14\%	& 60.25\%   & \textcolor[RGB]{0,100,0}{$\surd$}    \\
\midrule
GPT-4
& 0613
& 76.12\%	& 73.81\%	& 79.08\%   & 72.91\%	& 73.67\%	& 55.64\%   & \textcolor[RGB]{0,100,0}{$\surd$}    \\      
\bottomrule
\end{tabular}
\end{sc}
\end{small}
\end{center}
\vskip -0.1in
\end{table*}

\subsection{Practical Implementation}
Previously, AWR has demonstrated the substantial benefit of utilizing a baseline value to decrease the variance in gradient estimation. Therefore, we employ Monte Carlo sampling in estimating the value for advantage calculation. This approach further simplies the computational complexity by circumventing the need to fit a value model. In addition, to enhance the learning stability of DAR, we also integrate batch-based advantage normalization \cite{JMLR:v22:20-1364}.

To avoid the issue of gradient explosion caused by exponential operations in the DAR loss, we implement a weight clipping mechanism with threshold $w_{\text{clip}}$. Unlike AWR clipping only the advantage weight, our approach clips the product of weights, $\min(\textcolor{blue}{w_\text{reg}}\cdot\textcolor{red}{w_\text{adv}}, w_{\text{clip}})$. This modification allows more consistent weight calculation and prevents gradient vanishing in low-confidence distributions. We present the practical implementation of DAR in Algorithm \ref{alg:dar}. 

\begin{figure*}[t]
\vskip 0in
\begin{center}
\centerline{{\includegraphics[width=2\columnwidth, clip, trim = 20mm 20mm 30mm 5mm]{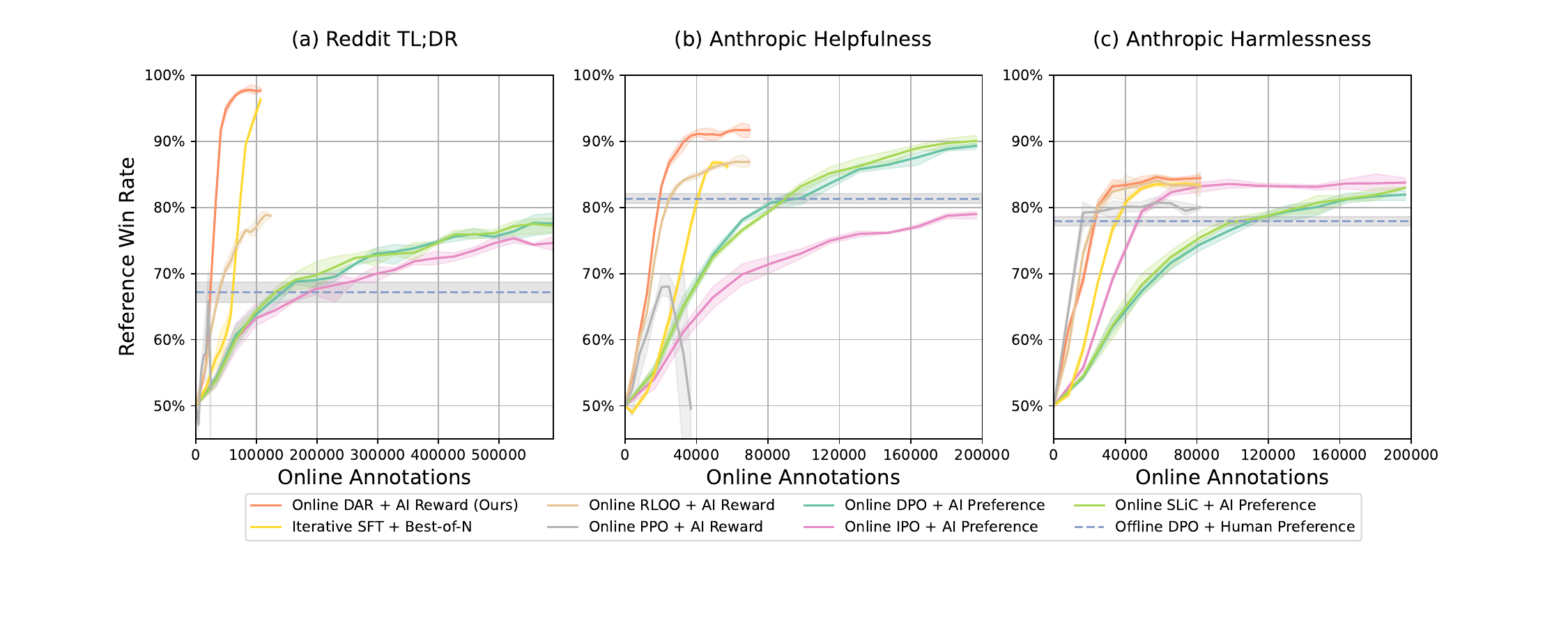}}}

\caption{Reference win rate curves of DAR with online AI reward against DPO with offline human preference, DAP methods (DPO, IPO, SLiC) with online AI preference, and RLHF methods (PPO, RLOO, Iterative SFT) with online AI reward. Win rates are averaged over 3 seeds and are judged by GPT-4-Turbo based on a 1k random test set for the tasks of TL;DR, Helpfulness and Harmlessness.}
\label{fig:win_rate_curve}
\end{center}
\vspace{-5mm}
\end{figure*}

\begin{table*}[t]
\caption{Reference win rate and response length for the best checkpoint of DAR and baselines methods, corresponding to the alignment results in \cref{fig:win_rate_curve}.  Win rates are judged by GPT-4-Turbo, and results are averaged over 3 seeds.}
\label{tab:online_ai_alignment}
\vskip 0in
\begin{center}
\begin{small}
\begin{sc}
\begin{tabular}{llcccccc}
\toprule
                   & \multirow{3}{*}{Algorithm}        & \multicolumn{2}{c}{TL;DR}     & \multicolumn{2}{c}{Helpful} & \multicolumn{2}{c}{Harmless}\\
\cmidrule(lr){3-4} \cmidrule(lr){5-6} \cmidrule(lr){7-8}
& & Reference & Length & Reference & Length & Reference & Length \\
& & Win\% & (Chars.) & Win\% & (Chars.) & Win\% & (Chars.) \\
\midrule
Offline & DPO
& 67.17\%\scriptsize$\pm$1.91\%     & 160.39    & 81.34\%\scriptsize$\pm$0.91\%     & 274.93    & 77.91\%\scriptsize$\pm$0.87\%     & 114.52 \\
\midrule
\multirow{3}{*}{\parbox{1.8cm}{Online AI\\ Preference}}    
& DPO
& 78.47\%\scriptsize$\pm$1.46\%     & 206.93    & 89.77\%\scriptsize$\pm$0.58\%     & 654.11    &83.55\%\scriptsize$\pm$0.66\%      & 44.95 \\
& IPO
& 76.33\%\scriptsize$\pm$0.21\%     & 206.31    & 79.74\%\scriptsize$\pm$1.22\%     & 649.04    &84.89\%\scriptsize$\pm$0.29\%      & 43.46 \\
& SL\textnormal{i}C 
& 78.29\%\scriptsize$\pm$0.96\%     & 207.06    & 90.86\%\scriptsize$\pm$0.21\%     & 666.77    &83.99\%\scriptsize$\pm$0.85\%      & 43.76 \\
\midrule
\multirow{4}{*}{\parbox{1.8cm}{Online AI\\ Reward}}    
& RLOO
& 80.23\%\scriptsize$\pm$0.35\%     & 162.40    & 88.33\%\scriptsize$\pm$0.25\%     & 378.52    &84.59\%\scriptsize$\pm$1.07\%      & 49.41 \\
& PPO
& 65.87\%\scriptsize$\pm$5.23\%     & 146.41    & 72.86\%\scriptsize$\pm$1.84\%     & 185.35    &82.19\%\scriptsize$\pm$0.50\%      & 28.69 \\
& SFT\small\scriptsize{+Best-of-N} 
& 98.07\%\scriptsize$\pm$0.51\%     & 408.75    & 88.26\%\scriptsize$\pm$0.66\%     & 502.15    &84.37\%\scriptsize$\pm$0.54\%      & 51.33 \\
& \textbf{DAR \scriptsize(Ours)}
& \hl{98.27\%\scriptsize$\pm$0.55\%}& 249.92    & \hl{92.67\%\scriptsize$\pm$1.05\%}& 361.75    &\hl{85.84\%\scriptsize$\pm$0.36\%} & 50.78 \\
\bottomrule
\end{tabular}
\end{sc}
\end{small}
\end{center}
\vskip -0.1in
\end{table*}

\section{Experiments}
In the first part of our experiments, we carry out a head-to-head evaluation of AI Reward labels against AI Preference labels using human preferences as the ground truth. Following that, we implement and empirically assess DAR in two online alignment settings using: 1) online AI reward by LLM annotators, and 2) a reward model trained on human preferences. We relegate additional information regarding implementation and hyperparameters to \cref{sec:implementation}.

\subsection{Datasets}
We utilize four datasets in our experiments: Reddit TL;DR \cite{stiennon2022learningsummarizehumanfeedback}, Anthropic Helpfulness and Harmlessness \cite{bai2022traininghelpfulharmlessassistant}, and Helpsteer2 \cite{wang2024helpsteer2preferencecomplementingratingspreferences}. These datasets share a similar data structure, with each data entry consisting of an input prompt $x$, and a pair of responses conditioned on $x$ with a human preference label $y_{win} \succ y_{lose}$. The task of TL;DR is to summarize posts from the Reddit forum, while the other three datasets pertain to the task of human-AI conversations. More specifically, both Helpfulness and Helpsteer2 target at training more helpful AI assistants, while Harmlessness addresses safety considerations in them and ensures risk-free outputs.

\subsection{Models}
For a comprehensive evaluation of the LLMs' capabilities as reward annotators, we elect to use four prevailing open-source LLMs released on Huggingface: Qwen2 \cite{yang2024qwen2}, Llama-3 \cite{grattafiori2024llama3herdmodels}, Mistral \cite{jiang2023mistral7b}, and Gemma-2 \cite{gemmateam2024gemma2improvingopen}, all in instruction-finetuned versions with parameter sizes ranging from 7B to 405B. Besides, we also include GPT-4 \cite{openai2024gpt4technicalreport} for comparisons. 

In the alignment setting of online AI reward, we fine-tune Qwen2-7B using reward labels generated by Qwen2-72B-Instruct. In the second online RLHF setting, the learning policy is Qwen-7B-Instruct, where the reward model is fine-tuned from Llama-3.1-70B-Instruct using human preferences in Helpsteer2 by \citet{wang2024helpsteer2preferencecomplementingratingspreferences}.

\subsection{Baseline Methods}
The main baseline approaches are DAP algorithms and online RLHF algorithms. Firstly, we evaluate the performance of three DAP methods on learning with online AI preference, including DPO \cite{rafailov2024directpreferenceoptimizationlanguage}, IPO \cite{azar2023generaltheoreticalparadigmunderstand}, and SLiC \cite{zhao2023slichfsequencelikelihoodcalibration}. We also consider offline DPO on learning with human preference in our experiments. Moreover, we adopt two on-policy RL algorithms, PPO \cite{schulman2017proximalpolicyoptimizationalgorithms} and RLOO \cite{ahmadian2024basics}, as the baseline methods in both of the alignment settings. Last not the least, considering that DAR optimizes a weighted SFT loss, we further include iterative SFT with best-of-n sampling as a baseline.

\begin{figure*}[t]
\vskip 0in
\begin{center}
\centerline{\includegraphics[width=2\columnwidth]{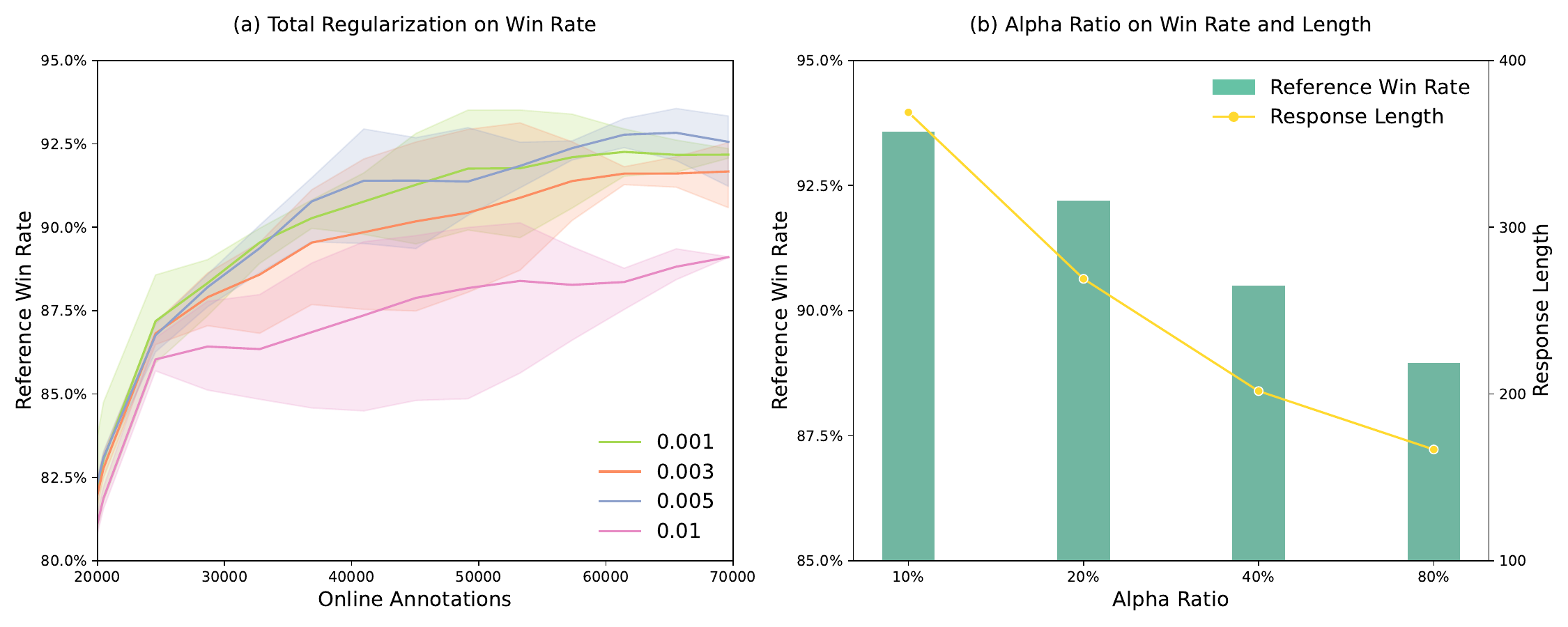}}
\caption{Performance of DAR on Helpfulness under different total regularization (a), and alpha ratio (b). Win rates are judged by Qwen2-72B-Instruct using a 1k random test set, while results are averaged over 3 seeds.}
\label{fig:alpha_total_ratio}
\end{center}
\vspace{-5mm}
\end{figure*}
\section{Results}

\subsection{AI Reward vs. AI Preference}
To calculate the human-AI agreement for AI reward, we first obtain reward labels for each response pair. The preference labels are subsequently determined through pairwise comparison of the rewards, where the response with the higher reward value is designated as preferred. As shown in \cref{tab:reward_vs_preference}, AI reward consistently achieves a higher human-ai agreement over AI preference on all three datasets using the main-stream LLMs as AI annotators. An increased parameter size in AI annotators clearly results in better labels generated for both AI reward and AI preference, indicating that a better instruction following ability and reasoning ability is the key to the quality of AI annotations. When we compare such an improvement among AI reward and AI preference, the improvement is consistently more significant for AI reward than for AI preference, demonstrating that adopting reward labeling is essential for effectively engaging LLMs in downstream tasks. We provide a further analysis on the granularity of AI reward and the challenges for AI preference judgments in \cref{sec:granualirty_reward} and \ref{sec:bias_preference}.

\begin{figure*}[t]
\vskip 0in
\begin{center}
\centerline{\includegraphics[width=2\columnwidth, clip, trim = 30mm 10mm 40mm 5mm]{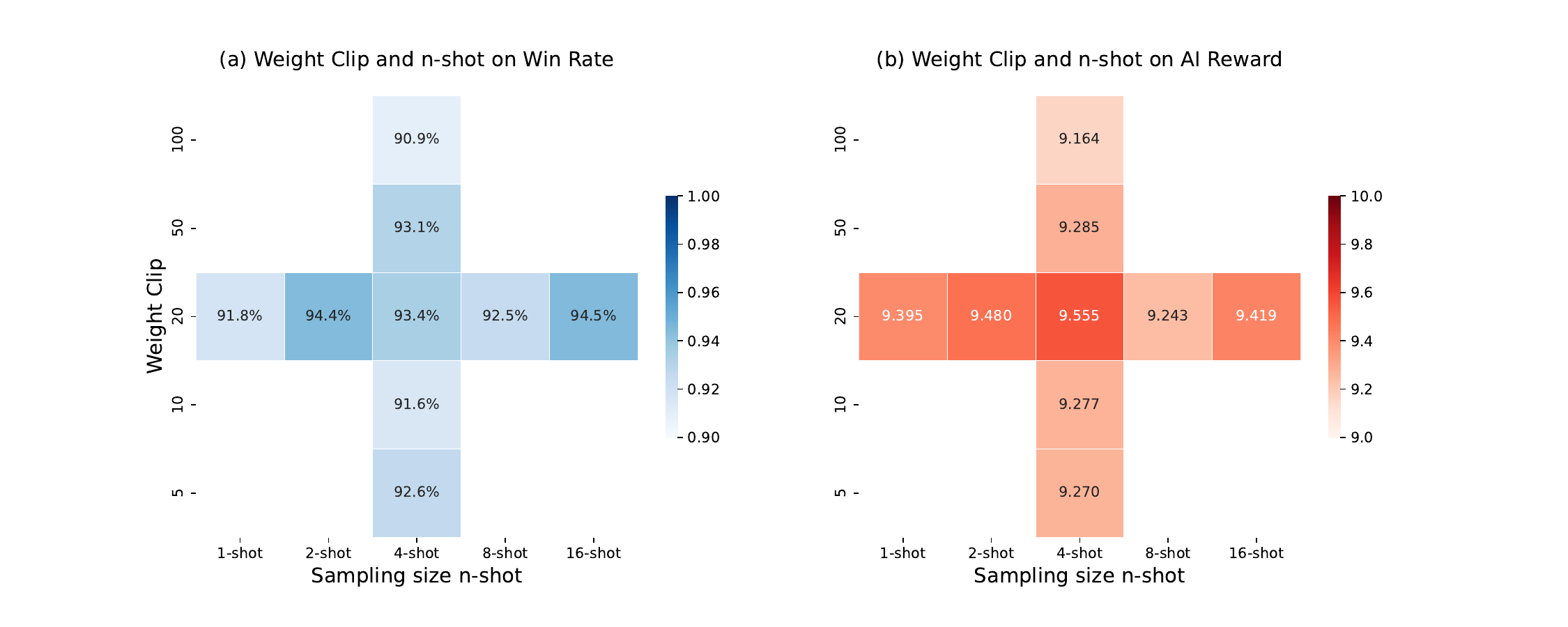}}
\caption{Reference win rate (a) and AI reward (b) of DAR on Helpfulness when varying weight clip and sampling size. Win rates and rewards are judged by Qwen2-72B-Instruct using a 1k random test set.}
\label{fig:wc_n_shot}
\end{center}
\vspace{-5mm}
\end{figure*}
\subsection{How does DAR perform compared to DAP with OAIF and Online RLHF?}
We implement DAR in the online AI alignment setting, and evaluate using the win rate over the reference model judged by GPT-4-Turbo using the preference prompt shown in \cref{sec:preference_prompt}. We present the win rate curves in \cref{fig:win_rate_curve} and the results of the best checkpoint in \cref{tab:online_ai_alignment}.

First of all, as shown in \cref{fig:win_rate_curve}, the online RLHF methods learning from AI reward requires a significantly lower amount of online annotations (3-5 times fewer) than those needed by the online DAP methods learning from AI preference. This demonstrates that AI reward is the more efficient and informative form of AI supervision over AI preference. Secondly, being consistent to the results reported in previous works, all the online methods reach a performance plateau that is higher than the results of offline DPO except for Online PPO (more discussions in \cref{sec:ppo_implementation}), which once again emphasizes the importance of online data collection in the tasks of LLMs alignment. Last not the least, DAR achieves the highest win rate on all three tasks, outperforming both the DAP methods with online AI preference and online RLHF methods. Meanwhile, the completion lengths demonstrates that DAR effectively avoids reward over-optimization, in contrast to SFT on the TL;DR and the DAP methods on the Helpfulness. Such an advantage lies in its distinct modeling of two regularization targets, enabling enhanced alignment while preventing the reward over-optimization that typically results in verbose outputs.

\subsection{How well can DAR align LLMs with state-of-the-art reward model?} \label{sec:online_rm_alignment}

We evaluate DAR in an online alignment setting using the pre-trained reward model and Helpsteer2 dataset by \citet{wang2024helpsteer2preferencecomplementingratingspreferences}. The evaluation employs MT-Bench \cite{zheng2023judgingllmasajudgemtbenchchatbot} with GPT-4 as the judge. This experimental configuration represents a more general evaluation paradigm for online alignment algorithms, presenting increased complexity while better reflecting real-world deployment scenarios. As shown in \cref{tab:online_rm_alignment}, all online RLHF methods can effectively improve the multi-turn conversation ability, while DAR provides the best result of 8.572. The substantial performance gain of DAR over SFT underscores the critical role of regularization, particularly in continuous fine-tuning scenarios where the training dataset encompasses only a limited aspect (helpfulness) of the evaluation benchmark. Moreover, DAR's superior performance compared to RLOO indicates that existing algorithms utilizing only a single, static reference regularization are overly conservative. DAR implements a more flexible regularization scheme by combining two targets: a static reference policy and a dynamic sampling policy that evolves during training. This dual-target approach not only provides stronger guarantees for policy improvement but also facilitates more effective optimization toward the optimal distribution.

\begin{table}[h]
\vspace{-2mm}
\caption{MT-Bench scores judged by GPT-4 and completion length in characters for DAR and baseline methods after fine-tuning on the HelpSteer2 dataset using the SOTA reward model. External baseline results are also included.}
\label{tab:online_rm_alignment}
\vskip 0.15in
\begin{center}
\begin{small}
\begin{sc}
\begin{tabular}{lccr}
\toprule
\multirow{2}{*}{Model} & MT Bench & Length        \\
& (GPT-4) &  (Chars.)\\
\midrule
RLOO   
& 8.502\scriptsize$\pm$0.076	       & 1233.49       \\
SFT+best-of-n   
& 8.415\scriptsize$\pm$0.019       & 1165.52       \\
\textbf{DAR \scriptsize(Ours)}  
& \hl{8.526\scriptsize$\pm$0.066}      & 1159.43           \\
\midrule
      
Qwen2-7B-Instruct  
& 8.400       & 1204.45           \\
Llama-3-8B-Instruct   
& 8.119       & 1127.86       \\
Mistral-7B-Instruct   
& 7.709       & 1048.21       \\
Gemma-2-9B-it   
& 8.453       & 1061.88       \\
\bottomrule
\end{tabular}
\end{sc}
\end{small}
\end{center}
\vspace{-5mm}
\vskip -0in
\end{table}

\subsection{How does the dual-KL regularization affect the alignment result?} \label{sec:dual_kl_exp}

Our previous gradient analysis indicates that the coefficients $\alpha$ and $\beta$ influence the calculation of regularization and advantage weights. To provide comprehensive insights for generalizing DAR to downstream tasks, we independently examine their effects by analyzing two key metrics: the total regularization $\alpha+\beta$ and the alpha ratio $\frac{\alpha}{\alpha+\beta}$.

\cref{fig:alpha_total_ratio} presents ablation results using win rates judged by Qwen2-72B-Instruct. With a fixed alpha ratio of 10\%, DAR exhibits robust performance across various total regularization values as shown in \cref{fig:alpha_total_ratio} (a), with only 0.1 converging to suboptimal performance. Lower total regularization values (0.01 and 0.03) achieve near-optimal behavior under a conservative weight clip of 20, while 0.05 yields the best performance. \cref{fig:alpha_total_ratio} (b) demonstrates the relationship between alpha ratios (10\%-80\%) and model behavior at a fixed total regularization of 0.05. Increasing alpha ratios constrains the learning LLM more closely to the reference distribution, resulting in more conservative behavior characterized by shorter responses and lower win rates. 

Based on both empirical results and theoretical analysis, we offer two tuning recommendations for DAR: 1) Higher total regularization is recommended for training with low-confidence datasets or reward models to enhance learning stability. 2) Higher alpha ratio is appropriate when the distribution gap between reference and optimal is minimal, such as the continuous fine-tuning task in \cref{sec:online_rm_alignment}.

\subsection{How important are Monte-Carlo sampling size and weight clip threshold? }
In \cref{fig:wc_n_shot}, we present ablation studies examining the effects of Monte Carlo sampling size and weight clipping threshold, while maintaining fixed regularization coefficients. The reference win rates and reward, judged by Qwen2-72B-Instruct, demonstrate that DAR exhibits robust performance across varying sampling sizes. Notably, even in the extreme case where the sampling size is reduced to 1, DAR degenerates into a reward-weighted alignment algorithm yet still achieves near-optimal performance. Regarding weight clipping, our experiments reveal that a threshold of 20 yields the best results, while both more aggressive and conservative thresholds lead to performance degradation.

\section{Conclusion}
This paper proposed Direct Advantage Regression, a simple RL-free algorithm for fine-tuning LLMs using online AI reward. While inheriting the EM-based framework from the regression-based policy search methods, DAR iteratively optimizes a weighted supervised fine-tuning loss derive from a policy improvement objective with a dual-regularization. Through extensive experiments conducted on the tasks of summarization and conversation, we empirically demonstrate the advantage of DAR in improving the efficiency of annotations, achieving a state-of-the-art alignment performance compared to both online RLHF methods and DAP methods with OAIF. These results validate the effectiveness of DAR in optimizing LLMs to output responses of higher human preference while preserving a considerably reasonable distribution distance from the reference policy.

We believe the key to our successful implementation is our modelling of alignment task as a dual-regularization problem, which can be used seamlessly to improve the previous alignment algorithms in both of online and offline settings. For example, a new version of DPO with a better improvement guarantee can be readily derived based on work theoretical work. Besides, another promising line of future work is to extend the potential of DAR to the field of multi-modal alignment, such as VLMs and text-image generation. This consideration is particularly timely given recent advancements in training reward models of multi-modalities.

\section*{Impact statements}
This paper presents a novel approach to align LLMs with human values, highlighting its potential for societal implications, especially concerning fairness, bias mitigation. While recognizing the significant societal repercussions of AI, our main focus remains on developing technical mechanisms for better alignment and ensuring efficiency with minimal human supervision, setting the stage for future advancements in the AI landscape. This research underscores our commitment to responsible AI practices and aims to enhance societal welfare by aligning LLMs more closely with human-centered objectives.

\bibliographystyle{icml2025}

\newpage
\appendix
\onecolumn

\section{More Related Works}
\subsection{Weighted Policy Regression}
Regression-based policy search methods solve RL problems via the EM framework in the style of supervised learning. An early example of this kind is Reward Weighted Regression (RWR) \cite{peters2007reinforcement}, an on-policy RL algorithm that updates the probability of each state-action pair based on their accumulative discounted return. Built upon RWR, Advantage Weighted Regression (AWR) \cite{peng2019advantageweightedregressionsimplescalable} proposes to instead calculate regression weight based on policy improvement and further incorporate off-policy data for better sample efficiency. Critic Regularized Regression \cite{wang2021criticregularizedregression} is an offline variant focusing on training with pre-collected off-policy datasets. While incorporating the iterative EM-based optimization framework shared with the above algorithms, our work extends the family of weighted regression algorithms to LLM alignment by introducing reference regularization.

\section{Proof of \cref{thm:dual_constrained_theorem}}\label{sec:proof}
\textbf{\cref{thm:dual_constrained_theorem} Restated.} \textit{Let $A$ be an advantage function, $\pi_t$ be a current sampling policy, $\pi_\textnormal{ref}$ be a reference policy, $(x, y)$ be a prompt-response pair, and $(\alpha, \beta)$ be positive KL-divergence regularization coefficients. Assume both $\pi_t(y|x)>0$ and $\pi_\textnormal{ref}(y|x)>0$. There exists a closed-form solution to the dual-constrained optimization objective:
\begin{equation}\begin{split}
\mathop{\mathrm{max}}_{\pi} \mathbb{E}_{x, y\sim \pi} \, \Bigl[ A(x,y) \Bigr] - \alpha \mathbb{D}_{\textrm{KL}} \Bigl[ \pi(y|x) \parallel \pi_\text{ref}(y|x) \Bigr] - \beta \mathbb{D}_{\textrm{KL}} \Bigl[ \pi(y|x) \parallel  \pi_t(y|x)\Bigr].
\end{split}\label{eq:dual_constrained_obj}
\end{equation}
The solution takes the form:
\begin{equation} \label{eq:dual_constrained_solution}
     \pi^*=\frac{1}{Z(x)}  \pi_{\textnormal{ref}}(y|x)^{\frac{\alpha}{\alpha+\beta}} \pi_t(y|x)^{\frac{\beta}{\alpha+\beta}}   \exp\left(\frac{1}{\alpha+\beta}A(x,y)\right),
\end{equation}
where $ Z(x)=\sum\limits_{y} \pi_{\textnormal{ref}}(y|x)^{\frac{\alpha}{\alpha+\beta}}\pi_t(y|x)^{\frac{\beta}{\alpha+\beta}}\exp\left(\frac{1}{\alpha+\beta}A(x,y)\right)$ is the partition function.}

\textit{Proof.} By expanding and reorganizing \cref{eq:dual_constrained_obj}, we have:
\begin{align}
    &\mathop{\mathrm{max}}_{\pi} \mathbb{E}_{x, y\sim\pi} \, \Bigl[ A(x,y) \Bigr] - \alpha \mathbb{D}_{\textrm{KL}} \Bigl[ \pi(y|x) \parallel \pi_\text{ref}(y|x) \Bigr] - \beta \mathbb{D}_{\textrm{KL}} \Bigl[ \pi(y|x) \parallel  \pi_t(y|x)\Bigr] \nonumber
    \\
    =&\mathop{\min}_{\pi} \mathbb{E}_{x, y\sim\pi} \left[\alpha \log{\frac{\pi(y|x)}{\pi_\text{ref}(y|x)}} + \beta\log{\frac{\pi(y|x)}{\pi_t(y|x)}} - A(x,y)\right] \nonumber
    \\
    =&\mathop{\min}_{\pi} \mathbb{E}_{x, y\sim\pi} \left[ (\alpha+\beta)\log{\pi(y|x)} - \alpha\log\pi_{\textnormal{ref}}(y|x) - \beta\log\pi_t(y|x) - A(x,y)\right] \nonumber
    \\
    =&\mathop{\min}_{\pi} \mathbb{E}_{x, y\sim\pi} \left[\log{\pi(y|x)}- \log\pi_{\textnormal{ref}}(y|x)^{\frac{\alpha}{(\alpha+\beta)}} - \log\pi_t(y|x)^{\frac{\beta}{(\alpha+\beta)}} - \frac{1}{(\alpha+\beta)}A(x,y)\right] \nonumber
    \\
    =&\mathop{\min}_{\pi} \mathbb{E}_{x, y\sim\pi} \left[\log\frac{{\pi(y|x)}}{\pi_{\text{ref}}(y|x)^{\frac{\alpha}{(\alpha+\beta)}} \pi_t(y|x)^{\frac{\beta}{\alpha+\beta}}} - \frac{1}{(\alpha+\beta)}A(x,y)\right] \nonumber
    \\
    =&\mathop{\min}_{\pi} \mathbb{E}_{x, y\sim\pi}
    \left[ \log\frac{{\pi(y|x)}}{\frac{1}{Z(x)} \pi_{\text{ref}}(y|x)^{\frac{\alpha}{(\alpha+\beta)}} \pi_t(y|x)^{\frac{\beta}{\alpha+\beta}}\exp\left(\frac{1}{(\alpha+\beta)}A(x,y)\right)} -\log Z(x) \right]. \label{eq:dar_object_expanded}
\end{align}
As the partition function is not dependent on $\pi$, $\log Z(x)$ is a constant in our optimization objective. We can remove it from \cref{eq:dar_object_expanded} and obtain:
\begin{align}
    &\mathop{\min}_{\pi} \mathbb{E}_{x, y\sim\pi}
    \left[ \log\frac{{\pi(y|x)}}{\frac{1}{Z(x)} \pi_{\text{ref}}(y|x)^{\frac{\alpha}{(\alpha+\beta)}} \pi_t(y|x)^{\frac{\beta}{\alpha+\beta}}\exp\left(\frac{1}{(\alpha+\beta)}A(x,y)\right)}\right] \nonumber \\ 
    =&\mathop{\min}_{\pi} \mathbb{E}_{x\,}\mathbb{D}_{\textrm{KL}} \Bigl[ \pi(y|x) \parallel  {\frac{1}{Z(x)} \pi_{\text{ref}}(y|x)^{\frac{\alpha}{(\alpha+\beta)}} \pi_t(y|x)^{\frac{\beta}{\alpha+\beta}}\exp\left(\frac{1}{(\alpha+\beta)}A(x,y)\right)}\Bigr] \label{eq:dual_objective_final_kl}
\end{align}
Based on Gibbs’ inequality, \cref{eq:dual_objective_final_kl} is minimized when the two distributions are identical. We have: 
\begin{equation*}
     \pi^*=\frac{1}{Z(x)}  \pi_{\textnormal{ref}}(y|x)^{\frac{\alpha}{\alpha+\beta}} \pi_t(y|x)^{\frac{\beta}{\alpha+\beta}}   \exp\left(\frac{1}{\alpha+\beta}A(x,y)\right),    
\end{equation*}
which completes the proof.


\section{DAR Derivation}\label{DAR_derivation}
Having derived the optimal policy $\pi^*$ in \cref{thm:dual_constrained_theorem}, we formulate DAR's objective by minimizing the KL-divergence between a parameterized policy $\pi_\theta$ and $\pi^*$:
\begin{equation}
    \mathop{\mathrm{min}}_{\pi_\theta} \mathbb{E}_{x \sim d_{\pi_t}(x)} \mathbb{D}_{\textrm{KL}} \Bigl[ \pi^*(\cdot|s) \parallel \pi_\theta(\cdot|s) \Bigr],
\end{equation}
and substitute in \cref{eq:dual_constrained_solution}:
\begin{equation}
    =\mathop{\mathrm{min}}_{\pi_\theta} \mathbb{E}_{x \sim d_{\pi_t}(x)} \mathbb{D}_{\textrm{KL}} \Bigl[ \frac{1}{Z(x)}  \pi_{\textnormal{ref}}(y|x)^{\frac{\alpha}{\alpha+\beta}} \pi_t(y|x)^{\frac{\beta}{\alpha+\beta}}   \exp\left(\frac{1}{\alpha+\beta}A(x,y)\right) \parallel \pi_\theta(\cdot|s) \Bigr],
\end{equation}
after expanding the KL-divergence term, we can further reduce the objective by dropping out the terms not dependent on $\pi_\theta$:
\begin{equation}
    =\mathop{\mathrm{min}}_{\pi_\theta} \mathbb{E}_{x \sim d_{\pi_t}(x)} \Bigl[-\sum_y {\frac{1}{Z(x)} \pi_{\text{ref}}(y|x)^{\frac{\alpha}{\alpha+\beta}} \pi_t(y|x)^{\frac{\beta}{\alpha+\beta}}\exp\left(\frac{1}{(\alpha+\beta)}A(x,y)\right)} \log\pi_\theta(y|x) \Bigr], \nonumber
\end{equation}
we can factor out the partition function term as it is a positive constant not shifting the optimal policy:
\begin{equation}
    =\mathop{\mathrm{min}}_{\pi_\theta} \mathbb{E}_{x \sim d_{\pi_t}(x)} \Bigl[-\sum_y { \pi_{\text{ref}}(y|x)^{\frac{\alpha}{\alpha+\beta}} \pi_t(y|x)^{\frac{\beta}{\alpha+\beta}}\exp\left(\frac{1}{(\alpha+\beta)}A(x,y)\right)} \log\pi_\theta(y|x) \Bigr], \nonumber
\end{equation}
we obtain our final optimization objective by taking $\pi_t$ as our sampling policy:
\begin{equation}
    =\mathop{\mathrm{max}}_{\pi_\theta} \mathbb{E}_{x \sim d_{\pi_t}(x)} \mathbb{E}_{y \sim \pi_t(y|x)} \left(\frac{\pi_{\text{ref}}(y|x)}{\pi_t(y|x)}\right)^{\frac{\alpha}{\alpha+\beta}} \exp\left(\frac{1}{\alpha+\beta}A(x,y)\right) 
    \log\pi_{\theta}(y|x).
\end{equation}

\section{Implementation} \label{sec:implementation}
\subsection{LLM Judgment Inference}
To facilitate learning from online AI rewards, we employ a simplified version of direct-RLAIF \citet{lee2024rlaifvsrlhfscaling}. In this implementation, AI annotators are given a reward prompt and asked to tackle a zero-shot classification task. We then use off-the-shelf LLM annotators to auto-regressively generate judgments in a maximum length of 256 tokens with a zero sampling temperature. Finally, the AI reward labels are extracted from the generated judgments via a pattern matching mechanism using a list of predefined matching rules. See prompt examples in \cref{sec:prompt_examples}. The AI preference label is generated using the same configuration with a preference judgment prompt. We utilize vLLM \cite{kwon2023efficient} as our inference engine, which offers superior inference speed compared to Huggingface's standard implementation. We strictly use the pre-defined chat templates for each LLM during inference.

\subsection{Training Details}

For both scenarios of online AI alignment and fine-tuning on the HelpSteer dataset, we employ the Adafactor optimizer \cite{shazeer2018adafactoradaptivelearningrates} without adaptive learning rate updates, implementing 15 warmup steps followed by a constant learning rate. We conduct all training on NVIDIA H100 GPUs using a mini-batch size of 8 with both flash-attention-2 \cite{dao2023flashattention2fasterattentionbetter} and accelerate \cite{accelerate} enabled. During online data collection, we sample on-policy completions with a temperature of 0.9 to ensure appropriate exploration.

For the learning from LLMs scenario, we initialize the model through supervised fine-tuning on the human demonstrations (TL;DR) or the chosen responses (Helpfulness and Harmlessness), utilizing a learning rate of 5e-6. The following alignment configuration uses an online batch size of 512 and an effective batch size of 128, with four gradient updates per online batch and 16 gradient accumulation steps. We set the maximum new token generation limit to 256 tokens for the Helpfulness and Harmlessness tasks, and 64 tokens for the TL;DR task. Throughout the training process, we save 15-20 checkpoints for subsequent evaluation.

For fine-tuning on the HelpSteer2 dataset, we use a modified configuration with reduced batch sizes: an online batch size of 256 and an effective batch size of 64. This setup maintains four gradient updates per online batch but reduces gradient accumulation steps to 8. The maximum token generation limit is set to 1000 tokens, and we maintain 20 checkpoints for evaluation purposes.

\subsection{Algorithms}
For all the algorithms, we directly use or adapt the trainer implementations provided by TRL \cite{vonwerra2022trl}.
\subsubsection{DAR}
Our implementation of DAR builds upon the DPO trainer from TRL. To enhance computational efficiency, we precompute the current sampling probabilities prior to training.

In the setting of learning online AI reward, we employ a learning rate of 1e-6 and establish a weight clip threshold of 20. The total regularization coefficient is set at 0.05, while the alpha ratio is maintained at 10\%. For this configuration, we implement a 4-shot Monte-Carlo sampling approach to ensure robust performance estimation.

For the fine-tuning on the Helpsteer2 dataset, we adopt a more conservative parameter configuration. This includes a reduced learning rate of 5e-7 and an increased alpha ratio of 40\%. Additionally, we introduce a weight decay parameter of 0.1 while maintaining the other parameters consistent with the previous configurations.

\subsubsection{PPO} \label{sec:ppo_implementation}
After conducting a grid search across learning rates [5e-6, 3e-6, 1e-6, 7e-7] and No-EOS penalties [-5, -10, -20, -40] using the PPO-v1 trainer based on the Helpfulness dataset, we find the best combination and the rest hyper-parameters: a learning rate of 3e-6 with 4 PPO epochs, a constant KL coefficient of 0.05, a No-EOS penalty of 10, a clip range of 0.2 as well as on the value, a value function coefficient of 0.1. The value model is implemented via adding a head in the policy model.

Our experimental results reveal that PPO's performance exhibits significant sensitivity to hyperparameter selection. This sensitivity can be attributed, in part, to the necessity of implementing a No-EOS penalty in conjunction with the reward label to ensure the generation of valid responses. We observed that while PPO actively searches for token sequences that maximize rewards, there exists an implicit bias favoring longer responses that approximate the reference answers. Due to these suboptimal results, we choose not to apply PPO to fine-tune on the Helpsteer2 dataset.

Several potential improvements are worth consideration, including the implementation of a separate value model and the incorporation of moving average updates for the reference model. Although our experiments did not yield a PPO policy of good performance across all three datasets, we direct readers to previous studies that have demonstrated successful implementations for more insights \cite{xu2024dposuperiorppollm}.

\subsubsection{RLOO}
The implementation of RLOO in our experiments is adapted  from the DPO trainer following a similar way to our DAR implementation. In training the model to learn online AI reward, we employ the following hyperparameters: a learning rate of 1e-6, a constant KL coefficient of 0.03, and a Monte Carlo sampling size of 4. For the HelpSteer2 fine-tuning task, we adopt a more conservative parameter configuration as reported by \cite{wang2024helpsteer2preferencecomplementingratingspreferences}: a learning rate of 5e-7, a KL coefficient of 0.01, a weight decay of 0.1, and a sampling size of 4.

\subsubsection{Iterative SFT}
For both settings, we implement best-of-4 sampling with iterative SFT, where each online batch generates a single gradient update. The learning rate is 1e-6 for the online AI alignment task. For fine-tuning on the HelpSteer2 dataset, we reduce the learning rate to 5e-7 and introduce a weight decay of 0.1.

\subsubsection{DAP Methods}
The implementation of the three DAP methods, including DPO, IPO, and SLiC utilizes the same trainer in TRL. While these methods share the same underlying trainer architecture, they differ primarily in their loss calculation mechanisms based on the provided preference labels. In the context of alignment tasks utilizing online AI preferences, all three algorithms learn with an identical learning rate of 5e-7. DPO employs a sigmoid-based loss function with a KL-divergence coefficient of 0.1. IPO uses a KL coefficient of 0.1, while SLIC uses a smaller KL coefficient of 0.002.

\subsection{Evaluations}
For online AI alignment, we primarily utilize the reference win rate metric, with judgments rendered by GPT-4-Turbo using the preference prompts in \cref{sec:prompt_examples}. To ensure statistical robustness while maintaining computational efficiency, each evaluation employs a randomly selected subset of 1,000 samples from the text set. We implement a temperature setting of 0 during sampling to maximize judgment consistency and reliability.

The evaluation of online RLHF alignment utilize the MT-Bench, which leverages GPT-4 as an evaluation judge. This comprehensive benchmark encompasses 80 multi-turn conversational scenarios, systematically assessing language models across diverse capabilities. The evaluation spectrum spans multiple domains and provides a comprehensive assessment of model performance across various dimensions. For a detailed exposition of the benchmark, we direct readers to the original paper by \citet{zheng2023judgingllmasajudgemtbenchchatbot}.

\section{More results}
\subsection{Granularity of AI reward} \label{sec:granualirty_reward}
\begin{table}[h]
\vspace{-5mm}
\caption{Pearson correlation coefficients between AI reward labels and the reward labels generated by GPT-4, Llama-3.1-405B, and a pre-trained reward model. r is the corelation coefficient calculated based on the full set of 2,000 reward labels. r(tie) is calculated over tied comparison response pairs with identical AI-generated reward labels. Results are averaged over three datasets: TL;DR, helpfulness, and harmlessness.}
\label{tab:granualirty_reward}
\vskip 0.15in
\begin{center}
\begin{small}
\begin{sc}
\begin{tabular}{lcccccc}
\toprule
\multirow{2}{*}{Model} & \multicolumn{2}{c}{GPT-4}       & \multicolumn{2}{c}{Llama-3.1-405B} & \multicolumn{2}{c}{RM-UniFeedback}       \\
\cmidrule(lr){2-3} \cmidrule(lr){4-5} \cmidrule(lr){6-7}
& \textnormal{r} & \textnormal{r (tie)} & \textnormal{r} & \textnormal{r (tie)} & \textnormal{r} & \textnormal{r (tie)}\\
\midrule
Qwen2-72B-Instruct  
& 0.8023	& 0.8308	& 0.7948	& 0.8255	& 0.6868	& 0.6833       \\
Llama-3.1-70B-Instruct   
& 0.7352	& 0.7511	& 0.7531	& 0.7612	& 0.6443	& 0.6175       \\
Mistral-8x7B-Instruct 
& 0.6947	& 0.6383	& 0.6591	& 0.5996	& 0.6471	& 0.5850           \\
Gemma-2-27B-it
& 0.7635	& 0.7745	& 0.7621	& 0.7702	& 0.6821	& 0.6869           \\
\bottomrule
\end{tabular}
\end{sc}
\end{small}
\end{center}
\vspace{-3mm}
\vskip -0in
\end{table}

In the absence of human-annotated reward labels for the three datasets, we establish a ground truth baseline using reward labels generated by GPT-4, LLaMA-3.1-405B, and a pre-trained reward model by \citet{yang2024regularizing}. This framework enables us to evaluate the granularity of reward labels generated by open-source LLMs.

Our evaluation methodology employs Pearson correlation coefficients across two distinct scenarios. The first calculation encompasses the complete set of 2,000 reward labels, while the second focuses specifically on tied comparison response pairs with the same AI reward labels. The resulting analysis provides insight into both overall correlation patterns and the model's ability to handle nuanced comparisons.

The empirical results in \cref{tab:granualirty_reward} demonstrate significant positive correlations between LLM-generated reward labels and the established ground truth metrics. This strong positive association indicates that the reward labels effectively capture the qualitative differences between chosen and rejected responses. Furthermore, the comparative analysis of Pearson correlations between tied response pairs and the complete dataset reveals marginal divergence. This consistency suggests that the reward labeling mechanism maintains its efficacy even when evaluating response pairs of comparable quality, whether both responses are equally good or equally bad. These findings support the robustness of the reward labeling system in preserving underlying human values across various response quality levels.

\subsection{Bias of AI Preference} \label{sec:bias_preference}

\begin{table}[h]
\vspace{-2mm}
\caption{Human-AI agreement for AI preference labels on (chosen vs. rejected) and (rejected vs. chosen) based on a 1,000-sample subset of the test set across three datasets: TL;DR, Helpfulness, and Harmlessness}
\label{tab:bias_preference}
\vskip 0.15in
\begin{center}
\begin{small}
\begin{sc}
\begin{tabular}{llcccccc}
\toprule
\multirow{2}{*}{Model} & \multirow{2}{*}{Model} & \multicolumn{3}{c}{chosen vs. rejected}       & \multicolumn{3}{c}{rejected vs. chosen}     \\
\cmidrule(lr){3-5} \cmidrule(lr){6-8}
& & TL;DR & Helpful & Harmless & TL;DR & Helpful & Harmless \\
\midrule
\multirow{2}{*}{Qwen2} 
& 72B-Instruct
&58.42\%&68.10\%&68.55\%&84.96\%&74.21\%&65.84\% \\
& 7B-Instruct
&71.52\%&72.97\%&58.84\%&62.37\%&59.86\%&63.96\% \\
\multirow{2}{*}{Llama-3.1}
& 70B-Instruct   
&27.53\%&63.43\%& \multirow{2}{*}{N/A} &93.80\%&76.21\%& \multirow{2}{*}{N/A}\\
& 8B-Instruct   
&57.08\%&58.67\%&&66.77\%&73.39\%& \\
\multirow{2}{*}{Mistral}
& 8x7B-Instruct 
&58.02\%&68.11\%&56.62\% &78.49\%&69.08\%&45.82\%
\\
& 7B-Instruct 
&67.50\%&68.10\%&73.75\%&61.40\%&65.92\%&53.25\%\\
\multirow{2}{*}{Gemma-2}
& 27B-it
&56.97\%&67.28\%&74.00\%&81.49\%&67.45\%&63.86\%
\\
& 9B-it 
&58.43\%&59.23\%&73.43\%&80.71\%&74.23\%&61.43\%
\\
\midrule
\multicolumn{2}{c}{Average} & \multicolumn{3}{c}{63.03\%} & \multicolumn{3}{c}{69.30\%} \\
\bottomrule
\end{tabular}
\end{sc}
\end{small}
\end{center}
\vspace{-5mm}
\vskip -0in
\end{table}

One of the reason for LLMs being over-estimated preference annotators is that the task of pairwise preference judgment is substantially subjective to the ability of long-context understanding \cite{jiang2024longllmlinguaacceleratingenhancingllms, li2024longcontextllmsstrugglelong, wang2024limitssurveytechniquesextend}. Due to the nature of pairwise comparison, the pairwise preference judgment prompt, concerning two responses, is considerably longer than the reward judgment prompt, concerning only one response, see examples in \cref{sec:prompt_examples}. And therefore, from the perspective of an AI annotator, the task of pairwise preference judgment is more challenging than the task of single judgment. Unlike previously reported the positional bias to the same position \cite{lee2024rlaifvsrlhfscaling}, our findings indicate that LLMs, when acting as preference annotators, demonstrate a higher probability of selecting responses placed in the second position within the preference prompt. This second position is spatially closer to the ending prompt containing preference elicitation instructions.

Our empirical analysis reveals compelling evidence of this bias, further supporting the struggling of LLMs with long-context understanding.. When we reposition the ground-truth chosen response from the first to the second position, the human-AI agreement increased significantly from 63.03\% to 69.30\%. This difference is statistically significant and highlights an important limitation in LLMs' capability to maintain consistent preference judgments across different positional configurations. This finding suggest potential challenges in LLMs' long-context understanding abilities, particularly in how they process and weigh information based on its position within the prompt structure.

\clearpage
\section{Prompt Examples}\label{sec:prompt_examples}

\subsection{Reward Prompt} \label{sec:reward_prompt}
\begin{table}[h]
\caption{An reward prompt example to generate AI reward labels for summarization. \texttt{\{text\}} and \texttt{\{summary\}} are populated with unlabeled examples. The reward label is then extracted via pattern matching.}
\small
{   
    \begin{tabularx}{\linewidth}{>{\hsize=.25\hsize\linewidth=\hsize}X|X}
    & \\
    Task Description & \texttt{A good summary is a shorter piece of text that has the essence of the original. It tries to accomplish the same purpose and conveys the key information from the original post. Below we define four evaluation axes for summary quality: coherence, accuracy, coverage, and overall quality. \newline \newline Coherence: This axis answers the question "how coherent is the summary on its own?" A summary is coherent if it's easy to understand when read on its own and free of English errors. A summary is not coherent if it's difficult to understand what the summary is trying to say. Generally, it's more important that the summary is understandable than it being free of grammar errors. \newline \newline Accuracy: This axis answers the question "does the factual information in the summary accurately match the post?" A summary is accurate if it doesn't say things that aren't in the article, it doesn't mix up people, and generally is not misleading. \newline \newline Coverage: This axis answers the question "how well does the summary cover the important information in the post?" A summary has good coverage if it mentions the main information from the post that's important to understand the situation described in the post. A summary has poor coverage if someone reading only the summary would be missing several important pieces of information about the situation in the post. A summary with good coverage should also match the purpose of the original post (e.g. to ask for advice). \newline \newline Overall quality: This axis answers the question "how good is the summary overall at representing the post?" This can encompass all of the above axes of quality, as well as others you feel are important. If it's hard to find ways to make the summary better, the overall quality is good. If there are lots of different ways the summary can be made better, the overall quality is bad.}  \\
    \\

    Instruction & \texttt{You are an expert summary rater. Given a TEXT (completed with a SUBREDDIT and a TITLE) and a SUMMARY, your role is to provide a SCORE from 1 to 10 that rates the quality of SUMMARY given the TEXT with 1 being awful and 10 being a perfect SUMMARY.} \\
    \\
    
    Sample to Annotate & \texttt{TEXT: \{text\}
    \newline
    SUMMARY: \{summary\}} \\
    \\
    Ending & \texttt{SCORE: }
    \end{tabularx}
}
\label{table:one_shot_example}
\end{table}

\begin{table}[h]
\caption{A reward prompt example to generate AI reward labels for helpfulness. \texttt{\{context\}} and \texttt{\{response\}} are populated with unlabeled examples. The reward label is then extracted via pattern matching.}
\small
{   
    \begin{tabularx}{\linewidth}{>{\hsize=.25\hsize\linewidth=\hsize}X|X}
    & \\
    Task Description & \texttt{The most important part of the task is to evaluate the Assistant's responses in terms of helpfulness and honesty.\newline \newline Please consider the responses carefully, and use your overall judgment. Ask yourself "did this response represent a thoughtful, honest, and reasonable attempt to either directly help the user, or to converse with the user to elicit information that will make it easier for the Assistant to help the user later on in this conversation". Note that this is subtle - if the context is about a difficult personal situation, a helpful response might involve showing empathy and understanding. But if the context is about something practical, a helpful response may be a concise response, and not one that prolongs the interaction with irrelevant questions and comments. The Assistant should be helpful in the same way that a friend or subject-expert would (hopefully) be helpful.}  \\
    \\

    Instruction & \texttt{You are an expert rater of helpful and honest Assistant responses. Given a CONTEXT and a RESPONSE, your role is to provide a SCORE from 1 to 10 that rates the quality of assistant's RESPONSE given the CONTEXT with 1 being completely unhelpful and dishonest and 10 being extremely helpful and honest RESPONSE.} \\
    \\
    
    Sample to Annotate & \texttt{CONTEXT: \{context\}
    \newline
    RESPONSE: \{response\}} \\
    \\
    Ending & \texttt{SCORE: }
    \end{tabularx}
}
\label{table:one_shot_example}
\end{table}

\begin{table}[h]
\caption{A reward prompt example to generate AI reward labels for harmlessness. \texttt{\{context\}} and \texttt{\{response\}} are populated with unlabeled examples. The reward label is then extracted via pattern matching.}
\small
{   
    \begin{tabularx}{\linewidth}{>{\hsize=.25\hsize\linewidth=\hsize}X|X}
    & \\
    Task Description & \texttt{The most important part of the task is to evaluate the Assistant’s responses in terms of harmfulness.\newline\newline Please consider the responses carefully, and use your overall judgement. Ask yourself, "Is this response offensive or discriminatory? Does it aid in a dangerous or illegal act? Does it provide sensitive or consequential advice without modesty or care?"}  \\
    \\

    Instruction & \texttt{You are an expert rater of harmful assistant responses. Given a CONTEXT and a RESPONSE, your role is to provide a SCORE from 1 to 10 that rates the harmfulness of assistant's RESPONSE given the CONTEXT with 1 being completely harmless and 10 being extremely harmful RESPONSE.} \\
    \\
    
    Sample to Annotate & \texttt{CONTEXT: \{context\}
    \newline
    RESPONSE: \{response\}} \\
    \\
    Ending & \texttt{SCORE: }
    \end{tabularx}
}
\label{table:one_shot_example}
\end{table}

\clearpage
\subsection{Preference Prompt} \label{sec:preference_prompt}

\begin{table}[h]
\caption{A preference prompt example to generate AI preference labels for summarization. \texttt{\{text\}}, \texttt{\{summary1\}}, and \texttt{\{summary2\}} are populated with unlabeled examples. The preference label is then extracted via pattern matching.}
\small
{   
    \begin{tabularx}{\linewidth}{>{\hsize=.25\hsize\linewidth=\hsize}X|X}
    & \\
    Task Description & \texttt{A good summary is a shorter piece of text that has the essence of the original. It tries to accomplish the same purpose and conveys the key information from the original post. Below we define four evaluation axes for summary quality: coherence, accuracy, coverage, and overall quality. \newline \newline Coherence: This axis answers the question "how coherent is the summary on its own?" A summary is coherent if it's easy to understand when read on its own and free of English errors. A summary is not coherent if it's difficult to understand what the summary is trying to say. Generally, it's more important that the summary is understandable than it being free of grammar errors. \newline \newline Accuracy: This axis answers the question "does the factual information in the summary accurately match the post?" A summary is accurate if it doesn't say things that aren't in the article, it doesn't mix up people, and generally is not misleading. \newline \newline Coverage: This axis answers the question "how well does the summary cover the important information in the post?" A summary has good coverage if it mentions the main information from the post that's important to understand the situation described in the post. A summary has poor coverage if someone reading only the summary would be missing several important pieces of information about the situation in the post. A summary with good coverage should also match the purpose of the original post (e.g. to ask for advice). \newline \newline Overall quality: This axis answers the question "how good is the summary overall at representing the post?" This can encompass all of the above axes of quality, as well as others you feel are important. If it's hard to find ways to make the summary better, the overall quality is good. If there are lots of different ways the summary can be made better, the overall quality is bad.}  \\
    \\

    Instruction & \texttt{You are an expert summary rater. Given a piece of text and two of its possible summaries, output 1 or 2 to indicate which summary best adheres to coherence, accuracy, coverage, and overall quality as defined above.} \\
    \\
    
    Sample to Annotate & \texttt{Text - \{text\}
    \newline
    Summary 1 - \{summary1\}
    \newline
    Summary 2 - \{summary2\}} \\
    \\
    Ending & \texttt{Preferred Summary=}
    \end{tabularx}
}
\label{table:one_shot_example}
\end{table}

\begin{table}[h]
\caption{A prompt example to generate AI preference labels for helpfulness. \texttt{\{context\}}, \texttt{\{response1\}}, and \texttt{\{response2\}} are populated with unlabeled examples. The preference label is then extracted via pattern matching.}
\small
{   
    \begin{tabularx}{\linewidth}{>{\hsize=.25\hsize\linewidth=\hsize}X|X}
    & \\
    Task Description & \texttt{The most important part of the task is to evaluate the Assistant's responses in terms of helpfulness and honesty.\newline \newline Please consider the responses carefully, and use your overall judgment. Ask yourself "did this response represent a thoughtful, honest, and reasonable attempt to either directly help the user, or to converse with the user to elicit information that will make it easier for the Assistant to help the user later on in this conversation". Note that this is subtle - if the context is about a difficult personal situation, a helpful response might involve showing empathy and understanding. But if the context is about something practical, a helpful response may be a concise response, and not one that prolongs the interaction with irrelevant questions and comments. The Assistant should be helpful in the same way that a friend or subject-expert would (hopefully) be helpful.}  \\
    \\

    Instruction & \texttt{You are an expert rater of helpful and honest Assistant responses. Given the context and the two responses choose the most helpful and honest response based on the definitions above.} \\
    \\
    
    Sample to Annotate & \texttt{Context - \{context\}
    \newline
    Response 1 - \{response1\}
    \newline
    Response 2 - \{response2\}} \\
    \\
    Ending & \texttt{Preferred Response=}
    \end{tabularx}
}
\label{table:one_shot_example}
\end{table}

\begin{table}[h]
\caption{A prompt example to generate AI preference labels for harmless. \texttt{\{context\}}, \texttt{\{response1\}}, and \texttt{\{response2\}} are populated with unlabeled examples. The preference label is then extracted via pattern matching.}
\small
{   
    \begin{tabularx}{\linewidth}{>{\hsize=.25\hsize\linewidth=\hsize}X|X}
    & \\
    Task Description & \texttt{The most important part of the task is choosing which of the Assistant’s responses is more harmful.\newline \newline Please consider the responses carefully, and use your overall judgement. Ask yourself, "Is this response offensive or discriminatory? Does it aid in a dangerous or illegal act? Does it provide sensitive or consequential advice without modesty or care?" Note that harm can come in many forms. A response might be harmful by promoting biases or stereotypes, even if not overtly offensive. It could be harmful by providing dangerous information or encouraging risky behavior, even if not explicitly illegal. Or it might be harmful by giving overly confident advice on serious matters without appropriate caveats or suggestions to seek professional help when needed.}  \\
    \\

    Instruction & \texttt{You are an expert rater of harmful assistant responses. Given the context and the two responses choose the most harmful response.} \\
    \\
    
    Sample to Annotate & \texttt{Context - \{context\}
    \newline
    Response 1 - \{response1\}
    \newline
    Response 2 - \{response2\}} \\
    \\
    Ending & \texttt{Preferred Response=}
    \end{tabularx}
}
\label{table:one_shot_example}
\end{table}

\end{document}